\documentclass[10pt,twocolumn,letterpaper]{article}

\usepackage{iccv}
\usepackage{times}
\usepackage{epsfig}
\usepackage{graphicx}
\usepackage{amsmath}
\usepackage{amssymb}
\usepackage{graphicx,array,multirow,float}

\iccvfinalcopy 

\def\iccvPaperID{3964} 

\ificcvfinal\fi

\begin{document}
\setlength\extrarowheight{5pt} 
\newcolumntype{P}[1]{>{\centering\arraybackslash}p{#1}}

\title{Simultaneous multi-view instance detection with learned geometric soft-constraints}

\author{Ahmed Samy Nassar\textsuperscript{1,2}, {S{\'e}bastien Lef{\`e}vre}\textsuperscript{1},  
Jan D. Wegner\textsuperscript{2}\\
\textsuperscript{1}IRISA, Universit{\'e} Bretagne Sud\\
\textsuperscript{2}EcoVision Lab, Photogrammetry and Remote Sensing group, ETH Zurich\\
\{ahmed-samy-mohamed.nassar, sebastien.lefevre\}@irisa.fr, jan.wegner@geod.baug.ethz.ch
}

\maketitle
\ificcvfinal\thispagestyle{empty}\fi

\begin{abstract}
We propose to jointly learn multi-view geometry and warping between views of the same object instances for robust cross-view object detection. What makes multi-view object instance detection difficult are strong changes in viewpoint, lighting conditions, high similarity of neighbouring objects, and strong variability in scale. 
By turning object detection and instance re-identification in different views into a joint learning task, we are able to incorporate both image appearance and geometric soft constraints into a single, multi-view detection process that is learnable end-to-end. 
We validate our method on a new, large data set of street-level panoramas of urban objects and show superior performance compared to various baselines. Our contribution is threefold: a large-scale, publicly available data set for multi-view instance detection and re-identification; an annotation tool custom-tailored for multi-view instance detection; and a novel, holistic multi-view instance detection and re-identification method that jointly models geometry and appearance across views.
\end{abstract}

\section{Introduction}


We propose a method to simultaneously detect objects and re-identify instances across multiple different street-level images using noisy relative camera pose as weak supervision signal. Our method learns a joint distribution across camera pose and object instance warping between views. While object detection in single street-level panorama images is straightforward since the introduction of robust, deep learning-based approaches like Faster R-CNN~\cite{ren2015faster} for object detection or Mask R-CNN~\cite{HeGDG17} for instance segmentation, establishing instance correspondences across multiple views with this wide baseline setting is very challenging due to strong perspective change between views. 
%
Moreover, Google street-view panoramas, which are our core data in this paper, are stitched together from multiple individual photos leading to stitching artefacts in addition to motion-blur, rolling shutter effects etc. that are common for these type of mobile mapping imagery. This makes correspondence search via classical structure-from-motion methods like~\cite{agarwal2009,agarwal2010} impossible.  


Our core motivation is facilitating city maintenance using crowd-sourced images. In general, monitoring street-side objects in public spaces in cities is a labor-intensive and costly process in practice today that is mainly carried out via in situ surveys of field crews. One strategy that can complement greedy city surveillance and maintenance efforts is crowd-sourcing information through geo-located images like proposed for street trees~\cite{wegner2016cataloging, branson2018google, lefevre2017toward}. 
We follow this line of work, but propose an entirely new simultaneous multi-view object instance detection and re-identification method that jointly reasons across multi-view geometry and object instance warping between views. We formulate this problem as an instance detection and re-identification task, where the typical warping function between multiple views of the same tree (Fig.~\ref{fig:Ng1}) in street-view panoramas is learned together with the geometric configuration. More precisely, instead of merely relying on image appearance for instance re-identification, we insert heading and geo-location of the different views to the learning process. Our model learns to correlate typical pose configurations with corresponding object instance warping functions to disentangle multiple possibly matching candidates in case of ambiguous image evidence. 

Our contributions are (i) a novel multi-view object instance detection and re-identification method that jointly reasons across camera poses and object instances, (ii) a new object instance re-identification data set with thousands of geo-coded trees, and (iii) a new interactive, semi-supervised multi-view instance labeling tool. We show that learning geometry and appearance jointly end-to-end significantly helps improving object detections across multiple views as well as final geo-coding of individual objects.

\section{Related Work}

We are not aware of any work that does simultaneous object detection and instance re-identification with soft geometric constraints. But our proposed method touches a lot of different research topics in computer vision like pose estimation, urban object detection, object geo-localization, and instance re-identification. A full review is beyond the scope of this paper and we thus provide only some example literature per topic and highlight differences with the proposed work.

\textbf{Pose estimation:} Learning to predict camera poses using deep learning has been made popular by the success of PoseNet~\cite{kendall2015posenet} using single RGB images and many works have been published since then~\cite{hideo2017,en2018,xiang2018}. Tightly coupling pose with image content can, for example, be helpful for estimating a human hand's appearance from any perspective if seen from only one viewpoint~\cite{poier2018learning}. Full human pose estimation is another task that benefits from combined pose reasoning across pose and scene content like~\cite{luvizon20182d} who employ a multi-task CNN to estimate pose and recognize action. In this paper, we rely on public imagery without fine-grained camera pose information.

\textbf{Urban object detection:} A large body of literature addresses urban object detection from an autonomous driving perspective with various existing public benchmarks, e.g. KITTI~\cite{geiger2013vision}, CityScapes~\cite{cordts2016cityscapes}, or Mapillary~\cite{neuhold2017mapillary}. In these scenarios, dense image sequences are acquired with minor viewpoint changes in driving direction with forward facing cameras. Such conditions make possible object detection and re-identification across views~\cite{chen2017multi, 8594049,zhao2018object}. 
In our setup, significant changes occur between views, thus making the re-identification problem much more challenging.


\textbf{Object geo-localization:} Geo-localization of objects from Google street-view imagery with noisy pose information was introduced in~\cite{wegner2016cataloging, branson2018google}. In a similar attempt, \cite{krylov2018automatic} geo-localize traffic lights and telegraph poles by applying monocular depth estimation using CNNs, then using a Markov Random Field model to perform object triangulation. The same authors extend their approach by adding LiDAR data for object segmentation, triangulation, and monocular depth estimation for traffic lights~\cite{krylov2018object}. \cite{zhang2018using} propose a CNN-based object detector for poles and apply a line-of-bearing method to estimate the geographic object position. 
We rather suggest here to follow an end-to-end learning strategy.

\textbf{Instance re-identification:} Matching image patches can be viewed as a simple version of re-identifying image content across different views, e.g. in structure-from-motion~\cite{han2015matchnet}, tracking~\cite{tao2016siamese}, super-resolution~\cite{yang2010exploiting}, depth map estimation~\cite{zbontar2016stereo}, object recognition~\cite{shechtman2007matching}, image retrieval~\cite{zheng2006effective}, and image classification~\cite{zhou2010image}. Our scenario is closely related to works on re-identifying object instances across views. Siamese CNNs have been established as a common technique to measure similarity, e.g. for the person re-id problem that tries to identify a person in multiple views~\cite{li2014deepreid}). 
\cite{xiao2017joint} detects and re-identifies objects in an end-to-end learnable CNN approach with an online instance matching loss. ~\cite{xiao2019ian} solves re-identification with a so-called center loss that tries to minimize the distance between candidate boxes in the feature space.  
In contrast to prior work~\cite{wegner2016cataloging, branson2018google, krylov2018object}, which does detection, geo-coding and re-identification in a hierarchical procedure, our method does it simultaneously in one pass. Methods based on Siamese models \cite{li2014deepreid} alone are not a viable solution to our problem, since they need image crops of the object and can not fully utilize re-identification annotations due to their pairwise labelling training setup. \cite{xiao2019ian} searches for a crop within the detections in a gallery of images, in comparison to our method which aims at matching detections from both full images. The key differences between our work and \cite{xiao2017joint} is that we both ensure object geolocalization and avoid storing features from all identities since that is impractical in a real-world application like the one considered in the paper where objects actually look very similar in appearance.



\begin{figure*}
\begin{center}
\includegraphics[width=\textwidth]{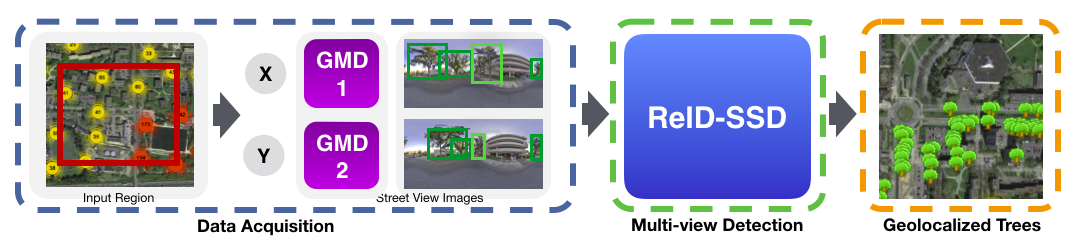}
\end{center}
   \caption{A pair of images is fed to our multi-view object detectors, matching projected predictions is learned, and the geo-coordinate of the object predicted.}
\label{fig:workflow}
\end{figure*}


\section{Multi-view detection and instance re-identification}

\begin{figure}
\vspace*{-0.2cm}
\centering
   \includegraphics[width=\columnwidth]{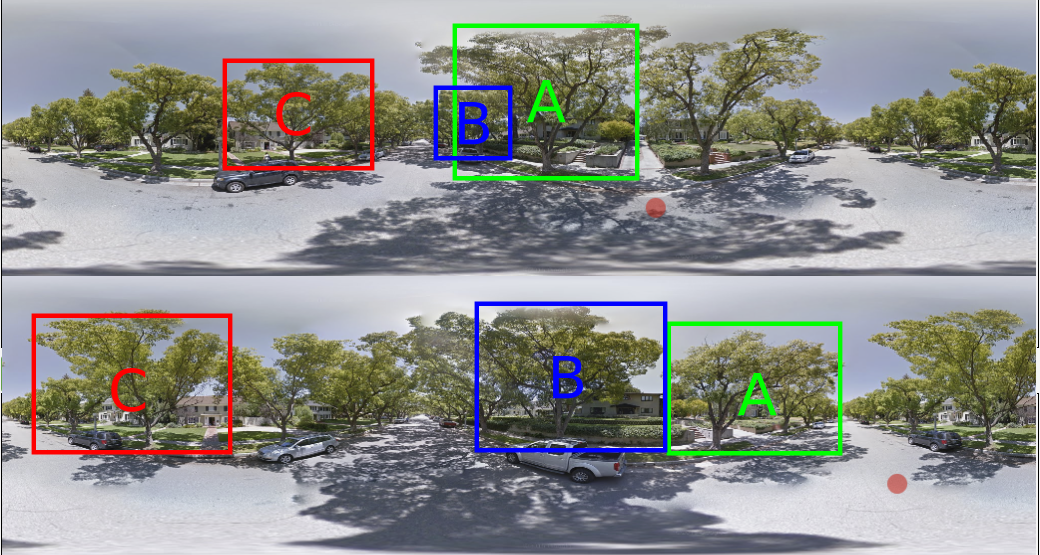}
   \caption{Tree instance re-identification problem (color indicates matches): each tree is photographed from multiple different views, changing its size, perspective, and background. Note that many trees look alike. }
   \label{fig:Ng2}
\vspace{-0.25cm}
\end{figure}

Our method learns to detect and re-identify object instances across different views simultaneously.  We compensate for inaccurate or missing image evidence by learning a joint distribution of multi-view camera poses together with the respective warping function of object instances.
Intuitively, our method learns to correlate a particular geometric pose setup (e.g., an equilateral triangle, a right triangle, etc.) with the corresponding change of object appearance in the images. As shown in Fig. \ref{fig:Ng2}, trees in many situations being the same species, and planted the same time look very similar making it hard to detect or re-identify. Learned relative camera pose configurations help re-identifying object instances across views if appearance information in the images is weak while strong image evidence helps improving noisy camera pose. In general, one can view the relative camera pose estimation task as imposing soft geometric constraints on the instance re-identification task. This joint reasoning of relative camera poses and object appearance ultimately improves object detection, instance re-identification, and also the final object geo-coding accuracy. 

\begin{figure}
\centering
   \includegraphics[width=150pt]{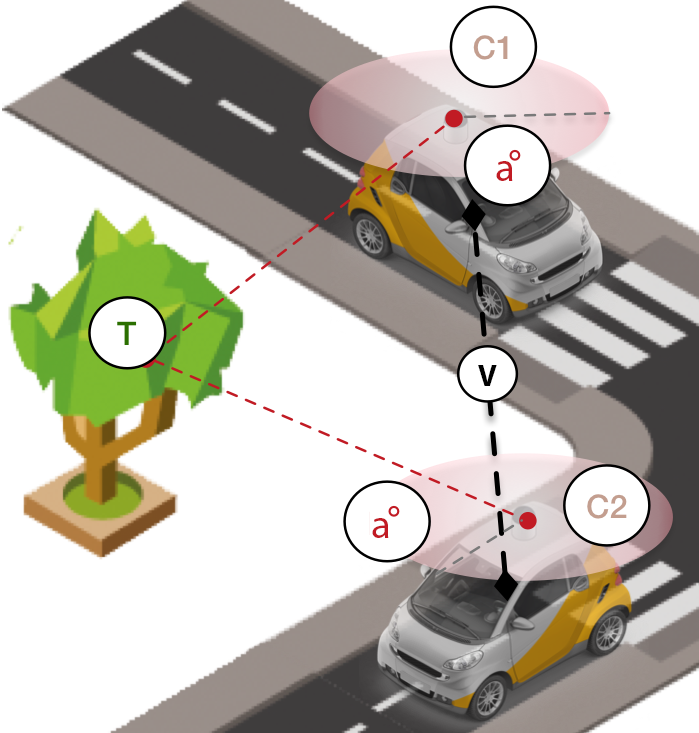}
   \caption{$C^*$: Camera with geo-position. $T$: The tree has its actual geographic coordinates, and location within the panorama. $a^\circ$: heading angle inside panorama. $v$: Distance between cameras.}
   \vspace{0.2cm}
   \label{fig:Ng1} 
\end{figure}

A big advantage of this simultaneous computation of relative camera poses, object instance warping and finally object geo-coding is that the model learns to compensate and distribute all small errors that may occur.
It thus implicitly learns to fix inaccurate relative poses relying on image evidence and vice versa. 
An overview of the architecture of our method is shown in Fig.~\ref{fig:mesh1}. The basic layout follows a Siamese architecture as proposed originally by~\cite{bromley1994signature}. The main concept of Siamese CNNs is constructing two identical network branches that share (at least partially) their weights. Features are computed for both input images and then compared to estimate the degree of similarity. This can be achieved by either evaluating a distance metric in the feature space or by evaluating the final classification loss. Here, our primary data source are Google street-view (GSV) panoramic images along with their geographic meta data (GMD) because they are publicly available at a global scale, fit our purpose of city-scale object mapping for maintenance purposes, and constructing large data sets amenable to deep learning is straightforward.  Fig. \ref{fig:Ng1} illustrates the setup of the problem, where the GSV panoramas captured from $C^*$ contain our object of interest $T$ from different viewpoints. The GMD contains many useful properties of the panorama image at hand but location in latitude and longitude as well as yaw are rather inaccurate. Since we do not have any information regarding $C$'s intrinsic or extrinsic properties, we rely on the GMD to use in our projection functions which plays an important role as we will present in the upcoming parts of the paper. GMD is also contains IDs of other images in vicinity.
%

\subsection{Multi-view object detection}
\label{mvod}
Our core object detection network component is based on the single shot detector (SSD)~\cite{liu2016ssd}. 
Our architecture is generally detector-agnostic and any detector could replace SSD if desired. We chose SSD over other prominent methods like Faster R-CNN~\cite{ren2015faster} because SSD provides an easy implementation that allows intuitive modifications and it performs faster with fewer classes, like in our case, while achieving good accuracy~\cite{huang2017speed}. We chose SSD512 \cite{liu2016ssd} as our preferred architecture, which sacrifices a bit of computational speed for better accuracy.

\begin{figure}[h]
    \centering
    \includegraphics[width=0.47\textwidth]{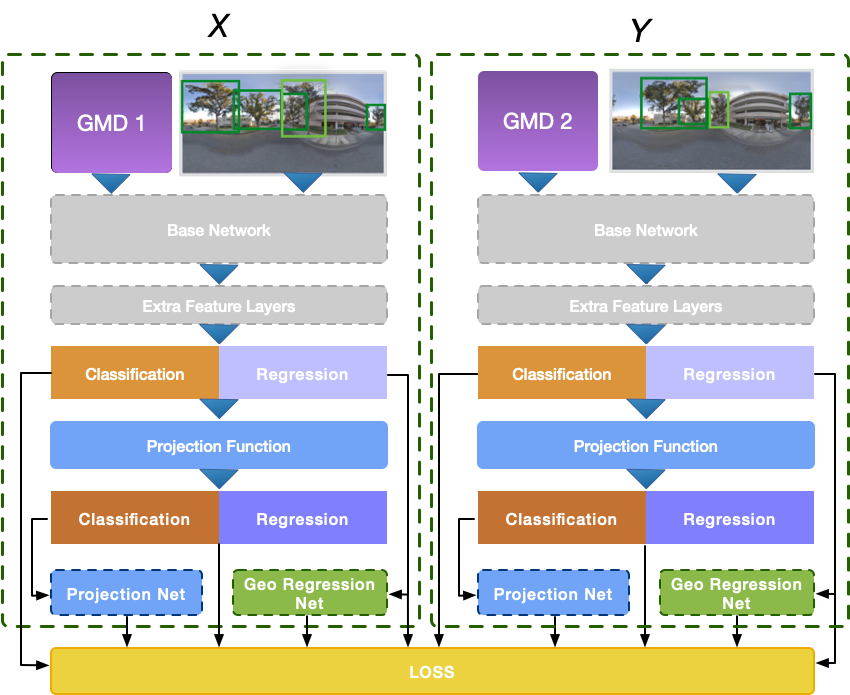}
    \caption{Our network design: Images along with their GMD are inputs to the network. Object bounding boxes and scores for each class are computed via extra feature layers (i.e., Conv4\_3 \cite{liu2016ssd}). These are projected into the other image's space (i.e. to the other network block, from $X$ to $Y$ and from $Y$ to $X$), and input to a dense \textit{Geo Regression Net} to estimate geographic coordinates. Finally, projected predictions are input to the \textit{Projection Net} network component that does some fine-tuning of the projections. }
    \label{fig:mesh1}
\end{figure}

Our network is composed of two identical blocks denoted as $ \textbf{\textit{X}} $ and $ \textbf{\textit{Y}} $ (Fig.~\ref{fig:workflow}). As shown in Fig.~\ref{fig:mesh1}, 
each block receives an image, camera pose information (geometric meta data, GMD), and the ground truth during training. Note that the camera's pose information $C$ only contains its location $\ell=(\mathrm{lat},\mathrm{lng})$, yaw, and height $ h $, which is passed to the network denoted as GMD in Fig.~\ref{fig:mesh1}. From the GMD data we are able to calculate the distance between the cameras, and the heading angle inside the panorama toward the object, see Eq.~\eqref{eq:streetview_enu}.
Ground truth is composed of two types of bounding boxes: (i) regular object bounding boxes and (ii) bounding boxes that carry instance IDs labeled and geo-coordinates. Each image passes first through the SSD base network composed of ResNet-50 modules~\cite{he2016deep}. It is then subject to the convolutional feature layers that provide us with detections at multiple scales. In order to prepare for instance re-identification, each individual object detection is given a \textit{local ID}, which will play a role in the multi-view instance matching stage later on. 
All detections per network block (i.e., panorama) are then projected during training into the other block's space using our geometric projection, see   Eq.~\eqref{eq:streetview_enu} \& \eqref{eq:streetview_geo2pix}. Predictions generated from $ \textbf{\textit{X}} $ and $ \textbf{\textit{Y}} $ are passed through a projection function that estimates their real world geographic position. From this position it is again projected into pixels into the corresponding view. These projection functions assume that the local terrain is flat to simplify the problem. Objects $ T $ are represented inside street view images in local East, North, Up (ENU)  coordinates that are calculated by providing $C_{l}$, $C_{h} $ and $ T_{l}$ using Eq.~\eqref{eq:streetview_enu}. To obtain the pixel location of the object $O_{x,y}$, Eq. \eqref{eq:streetview_geo2pix} is used given $R$ as the Earth's radius, $W$ and $H$ the image's width and height respectively, and $ z $ the estimated distance from $ C $ calculated by $z=\sqrt{e_x^2+e_y^2}$.

\begin{equation}
\label{eq:streetview_enu}
\begin{split}
(e_x,e_y,e_z) = \bigl( R \cos[\mathrm{C}_{lat}]\sin[\mathrm{T}_{lng}-\mathrm{C}_{lat}],\\ R\sin[\mathrm{T}_{lat}-\mathrm{C}_{lat}], -\mathrm{C}_{h} \bigr)
\end{split}
\end{equation}


\begin{equation}
\label{eq:streetview_geo2pix}
\begin{split}
 x =& \left(\pi + \arctan(e_x, e_y) - \mathrm{C}_{yaw}\right)W/{2\pi}\\
 y =& \left(\pi/2 - \arctan(-h, z) \right) H/{\pi}
\end{split}
\end{equation}



Blindly projecting bounding boxes between panoramas would, however, ignore any scale difference between different images of the same instance. Since the mapping vehicle is moving along while acquiring images, objects close to the camera in one image will likely be further away in the next. In addition, detected bounding boxes may sometimes be fitting an object inaccurately due to partial occlusions or simply poor detector performance. Using the above mentioned equations that assume flat terrain, these errors would results in projections meters away from the true position. We thus add a dense regression network to regress the predicted bounding boxes to the ground truth of the other block once projected. For example, $ \textbf{\textit{X}} $'s projected predictions are regressed to $ \textbf{\textit{Y}} $'s ground truth, and vice versa. This component (Geo Regression Net) aims at taking the predicted boxes, and projected boxes location, and regress them to their real world geo-coordinates through a densely layered network.

\paragraph{Geo Regression Net:}
Inspired by \cite{nilwong2018outdoor,sun2018accurate}, this network component estimates the geo-coordinates of the detected bounding boxes. Note that our ``Projection Function'' component (based on Eq.~\eqref{eq:streetview_enu} and~\eqref{eq:streetview_geo2pix}) provides initial estimates for geo-coordinates, which are improved with this component. The Geo Regression Net consists of two dense layers with ReLU activations.

\begin{figure}[h]
    \centering
    \includegraphics[width=0.48\textwidth]{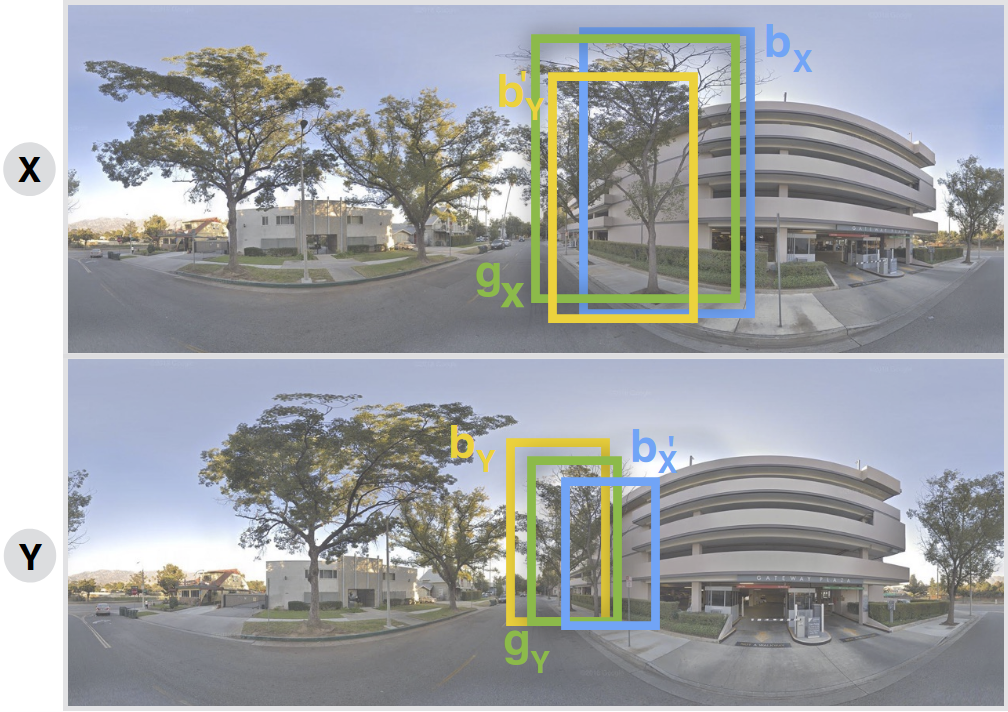}
    \caption{Illustration of predictions ($b_{*}$) projected ($b'_{*}$)  in other views and their ground truth ($g_{*}$).}
    \label{fig:projectionexplanation}
\end{figure}

\paragraph{Projection Net:}
This network component fine-tunes projected predictions $b'_*$ by learning to regress the discrepancy between them and the other block's ground truth as illustrated in Fig \ref{fig:projectionexplanation}. Projection Net is constructed similar to the extra feature layers (Fig.~\ref{fig:mesh1}), but uses only box regression layers. Our model was trained for 13 epochs, using a single NVIDIA GeForce GTX 1080Ti, with each epoch being trained for approximately 4.5 hours.


\subsection{Loss function}
We formulate a multi-task loss in Eq. \eqref{eq:loss_combined} to train our network. Similarly to SSD~\cite{liu2016ssd}, we use a softmax log loss $L_{conf}$ for classification and a smooth-L1 loss $L_{loc}$ for bounding box regression. 
As shown in Fig.~\ref{fig:projectionexplanation}, our predictions $b$ are projected into $b'$ using the projection function. We use again $L_{loc}$ but this time using the ground truth $g$ of the other image $g'$, since it contains the actual bounding boxes we are trying to regress to in order for the \textit{``Projection Network''} to regress the projected boxes.
However, mapping which predicted bounding boxes correspond to which default boxes $x$ in $g'$ is not a straightforward task due to how the default boxes are generated systematically: 
\begin{itemize}
    \item boxes inside $g$ and $g'$ are filtered ($f_g$ and $f_g'$) by keeping only the identified objects (ID'd) to ensure that we are regressing each instance to its corresponding box in the other image,
    \item $b$ is matched using IoU with $f_g$ to estimate which boxes are our target identities,
    \item indices of the boxes targeted are then selected to be used as inputs into our loss function, with $f_g'$ as ground truth.
\end{itemize}
The Geo Regression Net network component is trained using a RMSE $L_{RMSE}$ loss. For the re-identification task, we train both base networks $\textbf{\textit{X}}$ and $\textbf{\textit{Y}}$ using a contrastive loss $L_{cont}$ by feeding features from $x$ and $x'$ that are of identified objects as input to learn discriminative features and pull them close if similar. Our complete, multi-task loss function is:
\begin{equation}
\label{eq:loss_combined}
\begin{split}
L_{com}(x,c,b,g) = \frac{1}{N}(L_{conf}(x,c) + \alpha L_{loc}(x,b,g) \\
+ \alpha L_{loc'}(x,b',g,g') + L_{cont}(x,g)+L_{RMSE}(b,g))
\end{split}
\end{equation}

\label{sec:inference}
During inference, predicted boxes $b_*$ are combined in each view creating a large number of candidate boxes. As in the original implementation of SSD~\cite{liu2016ssd}, we use a classification confidence threshold of 0.01 to filter redundant boxes. Afterwards non-maximum suppression (NMS) with Jaccard index (IoU) is employed using a 0.5 overlap. As mentioned in Sec.~\ref{mvod} the \textit{local IDs} assigned are used to find which remaining candidate bounding boxes when projected, overlap's with the other view's candidate boxes (i.e. $b_{X} \cap b'_{Y}$), from which we can identify the corresponding boxes.
Simultaneously, by calculating the distance between the candidate boxes from each view using Euclidean distance, we are able to match corresponding boxes.

\section{Experiments}
We validate our method with experiments on two different data sets with street-level imagery. The first data set consists of GSV panoramas, meta data, and tree object instance labels across multiple views. The second data set contains sequences of Mapillary images acquired with dash cams where object instances are labeled across multiple consecutive image acquisitions. In addition to presenting final results of our end-to-end learnable multi-view object instance detection and re-identification approach, we also do a thorough ablation study to investigate the impact of each individual component. 

\subsection{Data sets}

\paragraph{Pasadena Multi-View ReID:} We build a new multi-view data set of street-trees, which is used as a test-bed to learn simultaneous object detection and instance re-identification with soft geometric constraints. The original Pasadena Urban Trees data set~\cite{wegner2016cataloging} contains 1,000 GSV images labeled using Mechanical Turk without explicit instance labels across multiple views. We construct a new \textit{Multi-View ReID} data set for our purpose where each tree appears in those four panoramas that are closest to a particular tree location. In total, we label 6,020 individual tree objects in 6,141 panoramas with each tree appearing in four different panoramas. Each panorama image is of size 2048 x 1024 px. This creates a total of 25,061 bounding boxes, where each box is assigned a particular tree ID to identify the different trees across panoramas. The annotations per image include the following: (i) bounding boxes identified (ID'd) and unidentified, (ii) ID'd bounding boxes include the geo coordinate position, distance from camera $v$, heading angle $a$, (iii) image's dimensions, and geo-coordinates. For validating our method experimentally, we split the data set into 4298 images for training, 921 for validation, and 922 for testing. Since we are not aware of any existing multi-view instance labeling tool for geo-located objects, we created a new one described in the sequel.

\paragraph{Mapillary:} In order to verify if our method generalizes across different data sets, we run experiments on a data set provided by Mapillary \footnote{www.mapillary.com}. Note that this particular data set is different from the well-known Mapillary Vistas dataset~\cite{neuhold2017mapillary}, which provides images and semantic segmentation. Our data set at hand is composed of 31,442 traffic signs identified in 74,320 images and carries instance IDs across views in an area of approximately 2$km^{2}$. On average, two traffic signs appear per image. 
This data set comes as GeoJSON ``FeatureCollection'' where each ``feature'' or identity contains the following properties that were used: (i) the object's geo-coordinate that is achieved by using 3D structure from motion techniques, therefore it is affected by the GPS, and the density of the images, (ii) the distance in meters from the camera position, (iii) image keys in which the object appears in and which is used to retrieve the image using their API, (iv) geo-coordinates of the image location, (v) the object's altitude, and (vi) an annotation in polygon form of the sign. 

The Mapillary data set significantly differs from our tree data set in several aspects. Images were crowd-sourced with forward looking dash cams attached to moving vehicles, and by walking humans using smart phones. Image sizes and image quality are thus inconsistent across the data set. Viewpoint changes between consecutive frames are only of a few meters, and the field of view per image is much smaller than a GSV panorama as shown in Fig.~\ref{fig:mapillary_sample}. Consequently, the distribution of relative poses between viewpoints is very different as well as the change in appearance of the same object instance across views. Because the camera is forward-looking, each object is viewed more or less from the same viewpoint, only scale changes. However, objects are generally smaller because unlike GSV, no orthogonal views perpendicular to the driving direction exist. 

\begin{figure}[!htbp]
\centering
\includegraphics[width=.22\textwidth]{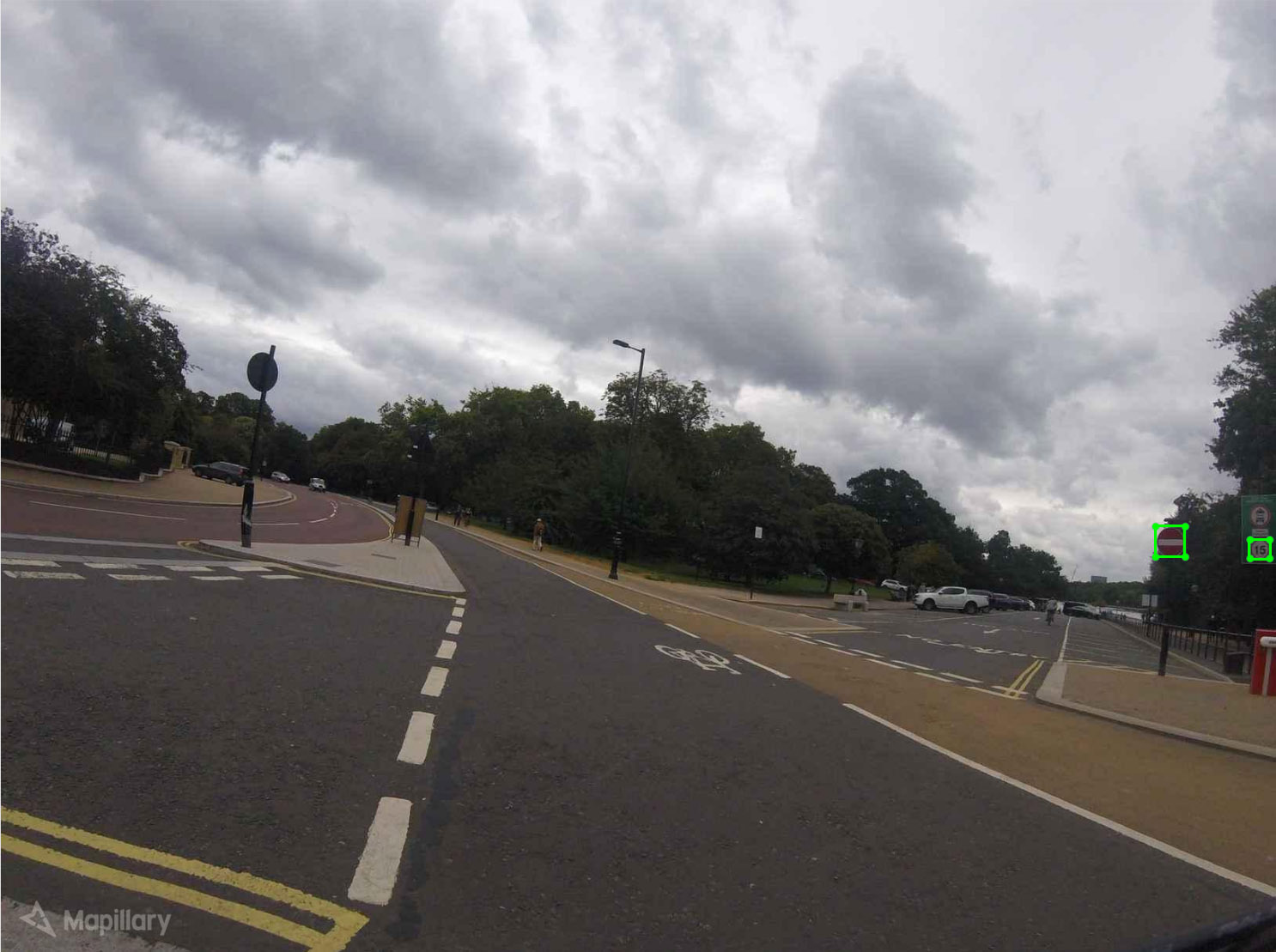}\quad
\includegraphics[width=.22\textwidth]{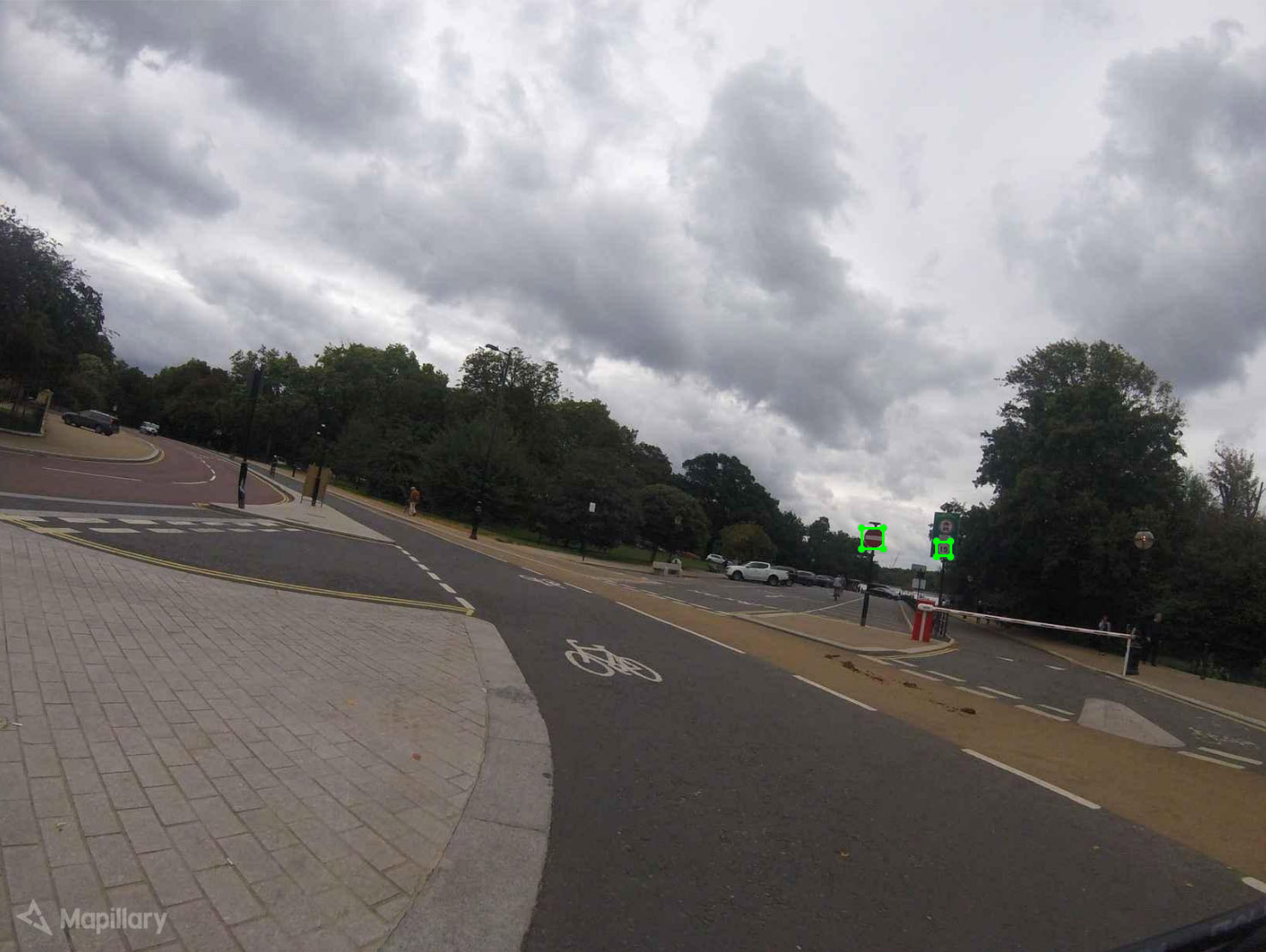}\quad

\medskip
\includegraphics[width=.22\textwidth]{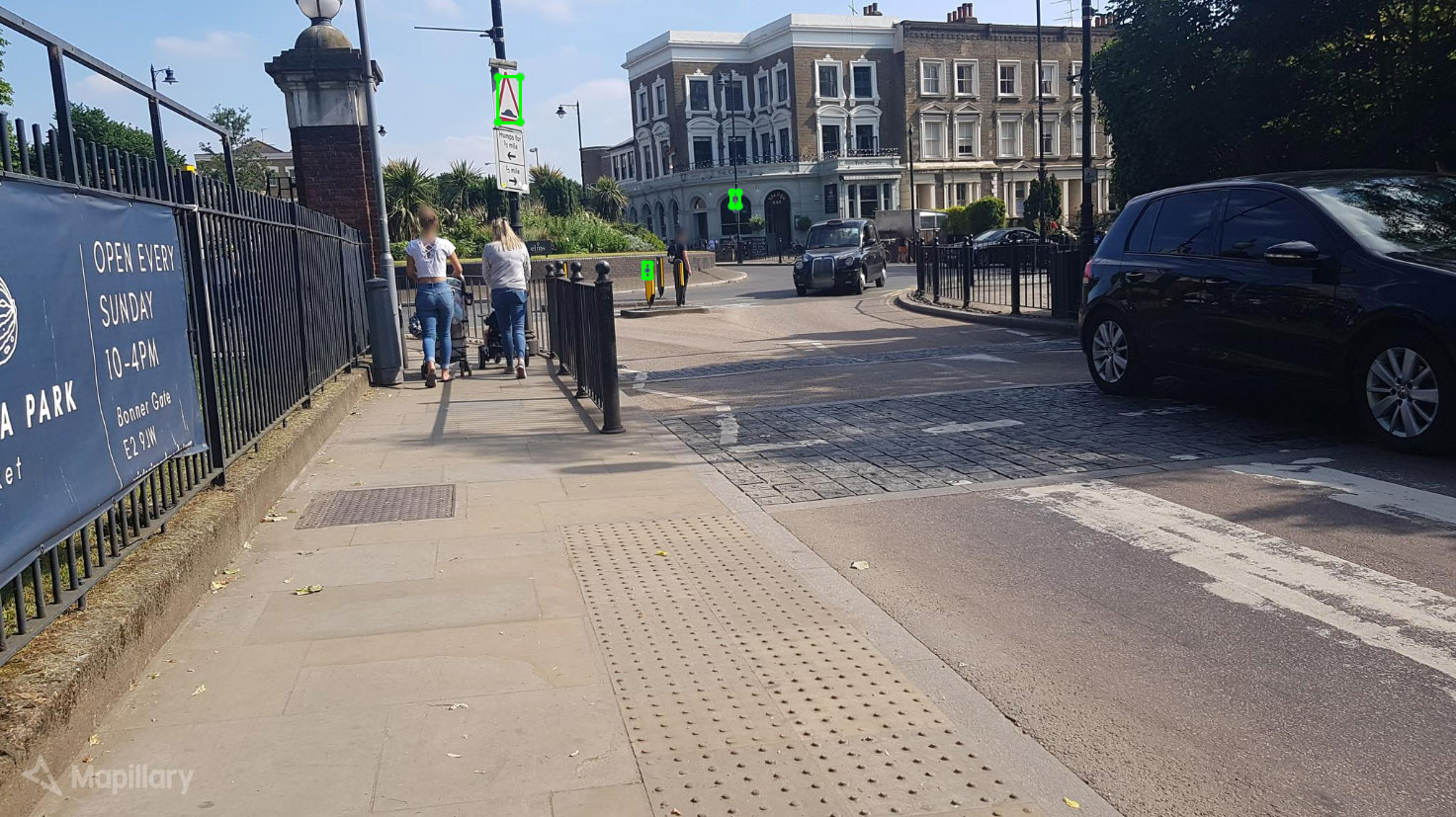}\quad
\includegraphics[width=.22\textwidth]{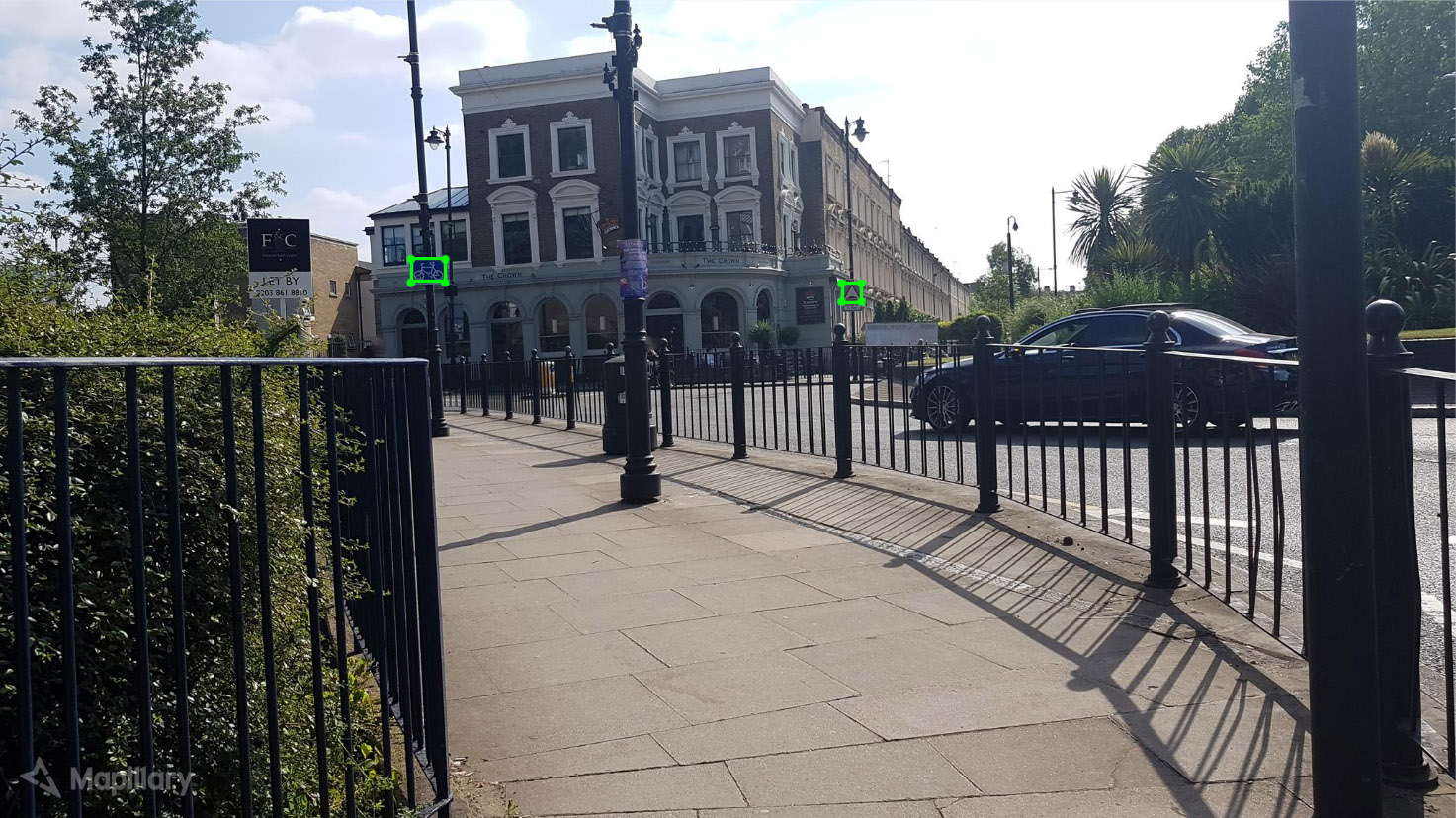}\quad

\caption{Consecutive frames of two example scenes of the Mapillary data set.}
\label{fig:mapillary_sample}
\end{figure}

\begin{figure}[!htbp]
\centering
\includegraphics[width=.5\textwidth]{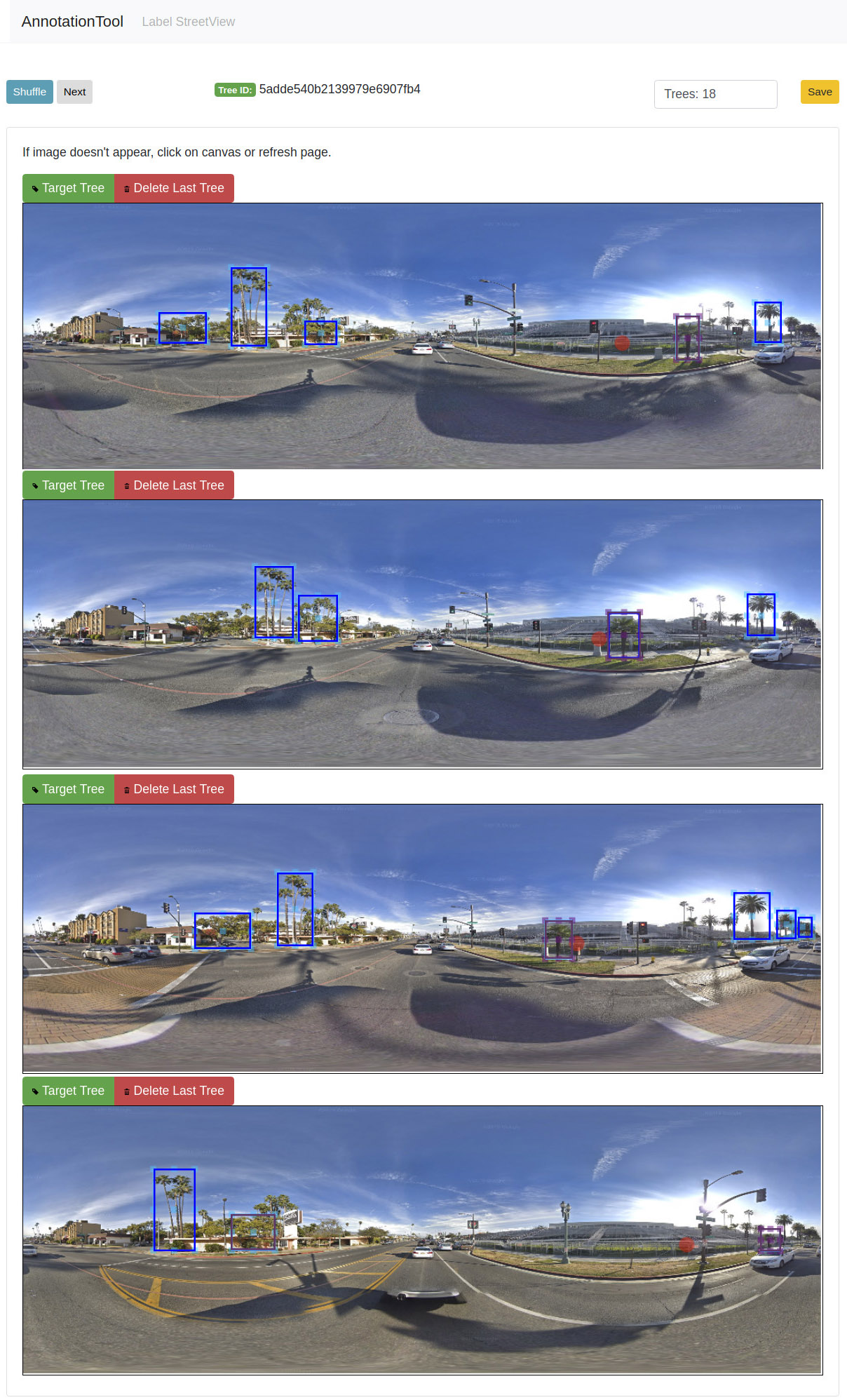}
\caption{Our annotation tool provides 4 multi-view panoramas from GSV. Initial bounding boxes for the target object are predicted, in which annotators can then refine, or annotate the missing object. To help identifying the object in multiple views a red circle is drawn to estimate the location of the target object in all views to guide the annotator. Images in the figure are from our Multi-View ReID dataset in Pasadena.}
\label{fig:refs}
\end{figure}

\subsection{Multi-view object annotation tool}
Labeling object instances across multiple panoramas is a difficult task (Fig.~\ref{fig:Ng2}) because many trees look alike and significant variations in scale and viewpoint occur. Our annotation tool aims at making multi-view instance labeling more efficient by starting from an aerial view of the scene. To begin labeling, the annotator first selects an individual object from the aerial image. The four closest panoramas are presented to the annotator and in each view a marker appears that roughly points at the object location inside each panorama. This projection from aerial view to street-view panorama approximates the object's position in each of the panoramas and is calculated using  Eq.~\eqref{eq:streetview_enu} and \eqref{eq:streetview_geo2pix}. This initial, approximate object re-identification significantly helps a human observer to identify the same tree across different images despite large scale and viewpoint changes. Moreover, SSD predicts bounding boxes around the objects of interest (here: trees) such that the annotator can simply refine or resize an existing bounding box in most cases (or create a new bounding box if the object remains undetected). Identity labeling or correspondence matching is done by selecting the best fitting bounding box (i.e., there may sometimes be more than one bounding box per object) per object per panorama. 
All multi-view instance annotations are stored in a MongoDB, which enables multiple annotators to work on the same data set at the same time. The database is designed to store annotated bounding boxes to each image, with a separate document storing which bounding boxes are identities. This reduces the effort of having to reannotate each image again for every identity.
%
%
Our labeling tool is generic in terms of object category and can easily be adapted to any category by re-training the detector component for a different class. Also the detector can be exchanged for any other object detector implemented/wrapped in Python. As output the labeling tool provides annotations through its API in VOC and JSON format. 




\subsection{Detection}
A significant benefit of projecting bounding boxes between blocks of our architecture is that it makes object detection much more robust against occlusions and missed detections caused by missing image evidence in individual images. 
In order to validate the improvement due to simultaneously detecting objects across multiple views, we compare object detector results on individual panoramas (Monocular) with results from our model that combine object evidence from multiple views via projecting detections between blocks $X$ and $Y$. Results are shown in Tab.~\ref{table:results_detection}. \emph{Ours} improves detection mAP on the Pasadena tree data set by 8.5 percent points, while improving by 2.7 percent points on the Mapillary data set. 
%
%

\subsection{Re-identification with pose information}
We verify if learning a joint distribution across camera poses and image evidence supports instance re-identification (regardless of the chosen architecture) with three popular Siamese architectures, namely FaceNet~\cite{schroff2015}, ResNet-50~\cite{he2016deep}, and MatchNet~\cite{han2015matchnet}. 
%
%
\begin{table}
\centering
\begin{tabular}{|c|c|c|} 
 \hline
 Method & w/o Pose [mAP] & w/ Pose [mAP]\\ [0.5ex] 
\hline
\hline
FaceNet~\cite{schroff2015}   & 0.808 & 0.842       \\ [0.8ex] 
ResNet-50~\cite{he2016deep} & 0.828 & 0.863       \\[0.8ex] 
MatchNet~\cite{han2015matchnet}  & 0.843 & \bf{0.871}  \\[0.8ex] 
\hline
\end{tabular}
\vspace*{3mm}

\caption{Re-identification results without (w/o Pose) and with (w/ Pose) camera pose information ($C^*_{l}$, $v$, $a$), fed to the Siamese network architectures FaceNet, ResNet-50, and MatchNet.}
\label{table:results_siamese}
\end{table}
Results shown in Tab.~\ref{table:results_siamese} indicate that \emph{Ours} with camera pose information consistently outperforms all baseline methods regardless of the base network architecture. Any architecture with added geometric cues does improve performance. Learning soft geometric constraints of typical scene configurations helps differentiating correct from wrong matches in intricate situations. Overall, \emph{Ours} with the MatchNet~\cite{han2015matchnet} architecture performs best. 

We evaluate the Re-ID mAP for our multi-view setting, which measures the amount of correct instance re-identifications if projecting detections between panoramas in the right column of Tab.~\ref{table:results_detection}. To measure the similarity between tree detections across multiple views projected onto one another, we use the distance between overlapping bounding boxes as explained in the inference stage. 73\% of all tree instances labeled with identities are matched correctly, which is a high number given the high similarity between neighboring trees and the strong variation in scale and perspective. As for Mapillary's dataset, 88\% of the traffic signs were re-identified. In comparison to tree objects, neighboring traffic signs (with different purposes) are easier to discriminate, but much smaller in size.

\subsection{Geo-localization}
\begin{figure}[!htbp]
\centering
\includegraphics[width=.5\textwidth]{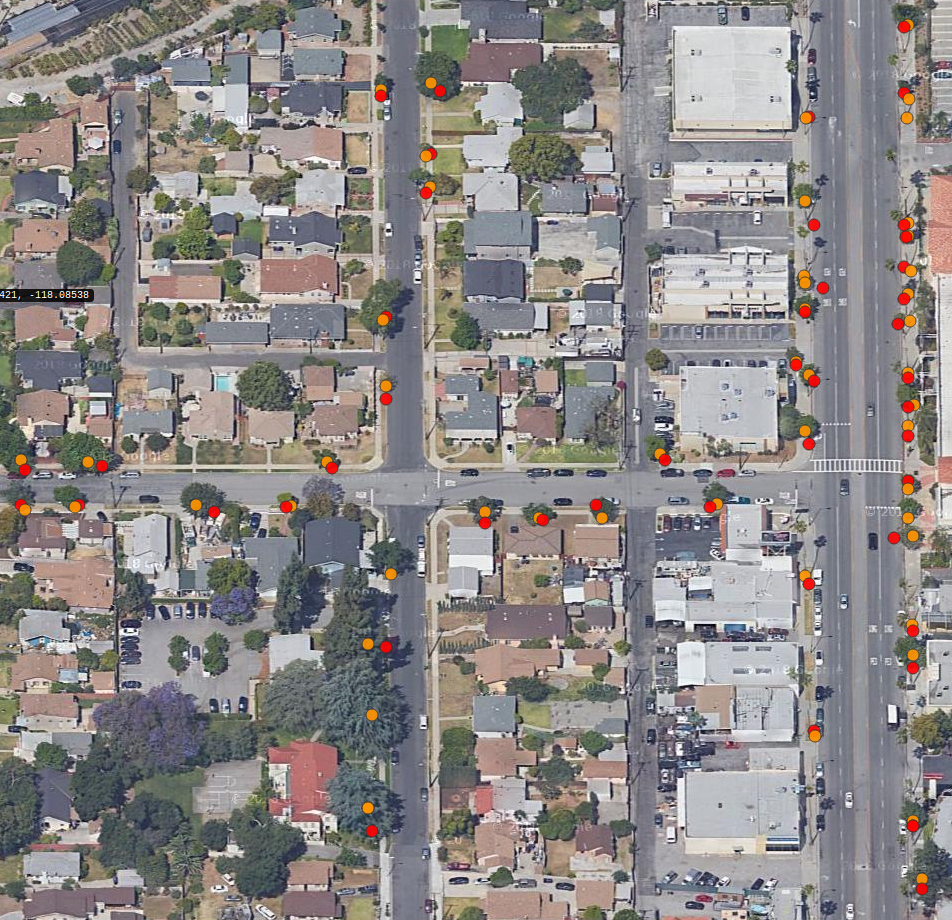}
\caption{Small subset of tree predictions (red) overlaid to an aerial image (not used in our model) and compared to tree ground truth locations (orange).}
\label{fig:aerial}
\end{figure}

We finally evaluate performance of our end-to-end trainable urban object mapping architecture by comparing predicted geo-locations of trees with ground truth positions. 
We compare our full, learned model (\emph{Ours}) against simply projecting each detection per single panorama (Single) to geographic coordinates as well as combining detections of multiple views (Multi) (Tab.~\ref{table:results_geolocalization}). We compute the discrepancy between predicted geo-coordinates and ground truth object position using the haversine formula given in  Eq.~\eqref{eq:haversine} with $r$ being the Earth's radius (6,372,800 meters):
\begin{multline}
\label{eq:haversine}
d = 2 r \arcsin\Biggl(\biggl(\sin^2\Bigl(\frac{b_{lat} - g_{lat}}{2}\Bigr) \\
+ \cos(g_{lat}) \cos(b_{lat})
\sin^2\Bigl(\frac{b_{lng} - g_{lng}}{2}\Bigr)\biggr)^{0.5} \Biggr)
\end{multline}
Single view geo-localization was done by applying projection functions given in Eq.~\eqref{eq:streetview_enu} and \eqref{eq:streetview_geo2pix} to the individual detections. As for multi-view experiments, we use combined detections from multiple views without learning the projection and project to geographic coordinates as before. \emph{Ours} is our full model as depicted in Fig.~\ref{fig:mesh1}, which takes advantage of ``Projection Net'' and ``Geo Regression Net'' components. Learning multi-view object detection and instance re-identification significantly improves performance, bringing down the MAE to 3.13 meters for the Pasadena trees data set while achieving 4.36 meters for Mapillary. Fig.~\ref{fig:aerial} shows tree detection results (red) for a small example scene in comparison to ground truth locations (orange) overlaid to an aerial view.  



\begin{table}[t]
\centering
\begin{tabular}{|c|c|c|c|} 
\hline
Method & Data set  & Det. mAP & Re-ID mAP\\ [0.5ex] 
\hline
\hline
\multirow{2}{*}{Monocular}
 & Pasadena  & 0.597 & - \\
 & Mapillary & 0.875 & - \\  [0.8ex] 
\hline\hline
\multirow{2}{*}{\emph{Ours}}
 & Pasadena & 0.682 & 0.731 \\ 
 & Mapillary & 0.902 & 0.882 \\[0.8ex] 
\hline
\end{tabular}
\vspace*{3mm}

\caption{Detection and Re-identification results with individual, single-view object detections (\emph{Monocular}) compared to our full, multi-view pipeline (\emph{Ours}).}
\label{table:results_detection}
\end{table}


\begin{table}[t]
\centering
\begin{tabular}{|c|c|c|} 
\hline
Method & Data set & MAE [m]\\ [0.5ex] 
\hline
\hline
\multirow{2}{*}{Single}
 & Pasadena  & 77.41 \\
 & Mapillary & 83.27 \\  [0.8ex] 
\hline
\hline
\multirow{2}{*}{Multi}
 & Pasadena  &  70.16 \\ 
 & Mapillary &   64.0 \\ [0.8ex]
 \hline
 \hline
 \multirow{2}{*}{\emph{Ours}}
 & Pasadena  &  \textbf{3.13}  \\ 
 & Mapillary &  \textbf{4.36} \\[0.8ex] 
\hline
\end{tabular}
\vspace*{3mm}

\caption{Geo-localization results as mean absolute error (MAE) compared to geographic ground truth object positions. }

\label{table:results_geolocalization}
\end{table}

\begin{figure}[!htbp]
\includegraphics[width=.5\textwidth]{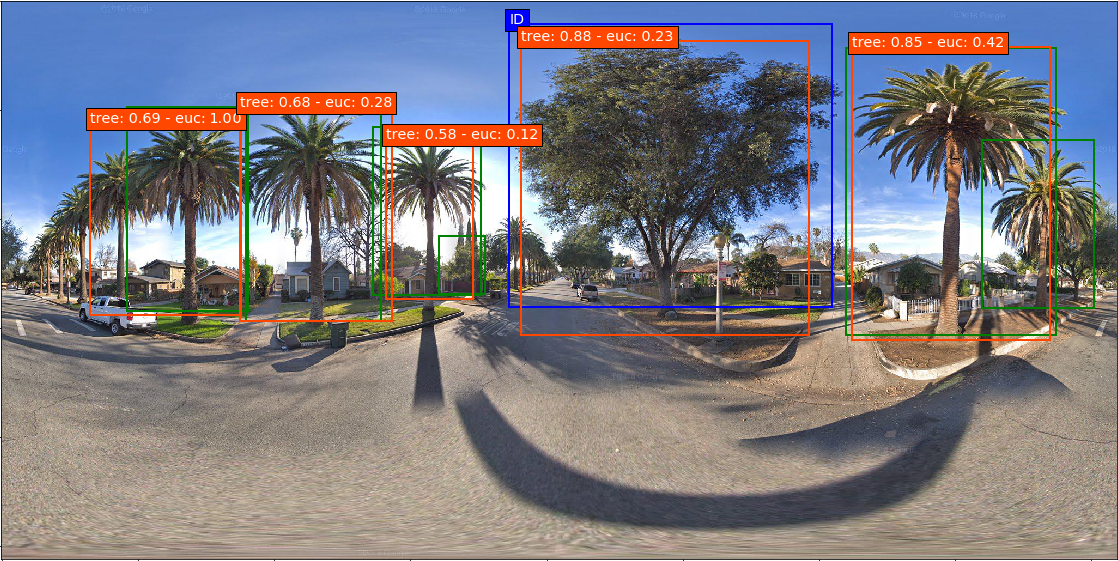}
\includegraphics[width=.5\textwidth]{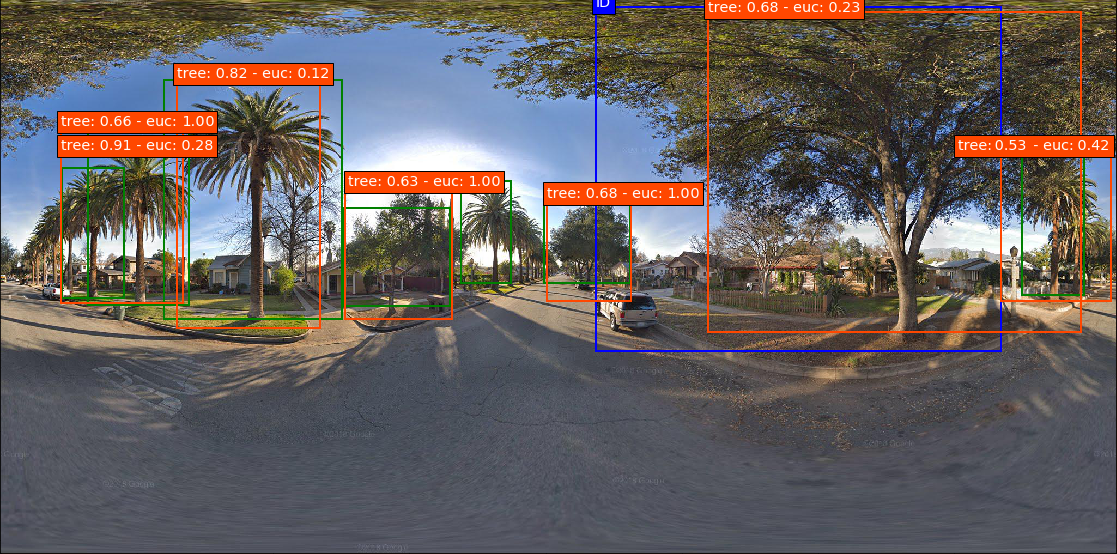}

\caption{Detection and Re-identification using our method. Green: ground truth boxes. Blue: ground truth box of the identity instance. Orange: predictions with classification score and calculated feature distance from matching box in the other view.}
\label{fig:mapillary_sample}
\end{figure}

%

\section{Conclusion}
We have presented a new, end-to-end trainable method for simultaneous multi-view instance detection and re-identification with learned geometric soft-constraints. Quantitative results on a new data set (labeled with a novel multi-view instance re-identification annotation tool) with street-level panorama images are very promising. Experiments on a Mapillary data set with shorter baselines, smaller objects, narrower field of view, and mostly forward looking cameras indicate that our method generalizes to a different acquisition design, too.

In general, integrating object evidence across views improves object detection and geo-localization simultaneously. In addition, our re-identification ablation study proves that learning a joint distribution across camera poses and object appearances helps re-identification. We hope tight coupling of camera pose information and object appearance within a single architecture will benefit further research on multi-view object detection and instance re-identification in the wild. All source code, the tree detection and re-identification data set, and our new labeling tool will be made publicly available\footnote{www.registree.ethz.ch}.

\textbf{Acknowledgements:} We thank Mapillary Research for providing the Mapillary dataset and the Hasler Foundation for funding our research presented here, which is part of the project "DeepGIS: Learning geographic information from multi-modal imagery and crowdsourcing".

\end{document}